\documentclass[times,twocolumn,final]{elsarticle}

\usepackage{arxiv}
\usepackage{graphicx}%
\usepackage{multirow}%
\usepackage{amsmath,amssymb,amsfonts}%

\usepackage{amsthm}%
\usepackage{mathrsfs}%
\usepackage[title]{appendix}%
\usepackage{xcolor}%
\usepackage[utf8]{inputenc}
\usepackage{booktabs}%
\usepackage{algpseudocode,algorithm}
\MakeRobust{\Call}
\usepackage{pifont}% http:ctan.org/pkg/pifont
\usepackage{adjustbox}
\usepackage{graphicx}
\usepackage{float}
\usepackage{placeins}
\usepackage{fancyhdr}
\usepackage[hyphens]{url}
\usepackage{hyperref}
\usepackage{caption}

\usepackage{subcaption}
\usepackage[font=small,labelfont=bf,labelsep=period,skip=4pt]{caption} % nicer caption
\usepackage{booktabs,tabularx,ragged2e,array,multirow,graphicx,longtable}
\newcolumntype{Y}{>{\RaggedRight\arraybackslash}X}

\fancypagestyle{firstpagestyle}{
    \fancyhf{} % Clear header and footer
    \fancyhead{} % Clear header
    \fancyfoot{} % Clear footer
}

\fancypagestyle{default}{
    \fancyhf{}
    \fancyhead[R]{\thepage} % Page number on the top right
}
\begin{document}

\title{Expert Consensus-based Video-Based Assessment Tool for Workflow Analysis in Minimally Invasive Colorectal Surgery: Development and Validation of ColoWorkflow}

\author[1,2]{Pooja P \snm{Jain}}
\author[1,3]{Pietro \snm{Mascagni}\corref{corresp}}
\cortext[corresp]{Corresponding author: \texttt{pietro.mascagni@ihu-strasbourg.eu}}
\author[1,3]{Giuseppe \snm{Massimiani}}
\author[2]{Nabani \snm{Banik}}
\author[4]{Marta \snm{Goglia}}
\author[2]{Lorenzo \snm{Arboit}}
\author[2]{Britty \snm{Baby}}
\author[5]{Andrea \snm{Balla}}
\author[6]{Ludovica \snm{Baldari}}
\author[4]{Gianfranco \snm{Silecchia}}
\author[3]{Claudio \snm{Fiorillo}}
\author[]{CompSurg Colorectal Experts Group}
\author[3]{Sergio \snm{Alfieri}}
\author[5]{Salvador \snm{Morales-Conde}}
\author[7,8]{Deborah S \snm{Keller}}
\author[6]{Luigi \snm{Boni}}
\author[1,2]{Nicolas \snm{Padoy}}

\address[1]{IHU Strasbourg, Strasbourg, France}
\address[2]{University of Strasbourg, CNRS, INSERM, ICube, UMR7357, Strasbourg, France}
\address[3]{Fondazione Policlinico Universitario Agostino Gemelli IRCCS, Rome, Italy}
\address[4]{Department of Medical-Surgical Sciences and Translational Medicine, Faculty of Medicine and Psychology, Sapienza University of Rome, Italy}
\address[5]{University Hospital Virgen Macarena, University of Sevilla, Seville, Spain}
\address[6]{Department of General and Minimally Invasive Surgery, Fondazione IRCCS Ca' Granda Ospedale Maggiore Policlinico, Milan, Italy}
\address[7]{Arizona State University, Tempe, AZ, USA}
\address[8]{Mayo Clinic, Phoenix, AZ, USA}

\received{XXX}
\finalform{XXX}
\accepted{XXX}
\availableonline{XXX}
\communicated{XXX}

\begin{abstract}

\textbf{Background:} Minimally invasive colorectal surgery is characterized by significant procedural variability, a difficult learning curve, and complications that impact patients' outcomes. Video-based assessment (VBA) offers an opportunity to generate data-driven insights to reduce variability, optimize training, and improve surgical performance. This study aims to develop and validate a VBA tool for workflow analysis across minimally invasive colorectal procedures. 

\textbf{Methods:} A three-round modified Delphi process was conducted among experts in colorectal surgery (CRS) and VBA to achieve consensus on generalizable workflow descriptors. The resulting framework informed the development of a new VBA tool, \textit{ColoWorkflow}. Independent raters then applied \textit{ColoWorkflow} to a multicentre video dataset of laparoscopic and robotic CRS. Applicability and inter-rater reliability were evaluated.

\textbf{Results:} Consensus was achieved for 10 procedure-agnostic phases and 34 procedure-specific steps describing CRS workflows. \textit{ColoWorkflow} was developed and applied to 54 colorectal operative videos (left and right hemicolectomies, sigmoid and rectosigmoid resections, and total proctocolectomies) from five centres in four countries. The tool demonstrated broad applicability, with all but one label utilized. Inter-rater reliability was moderate, with mean Cohen’s $\kappa$ of 0.71 for phases and 0.66 for steps. Most discrepancies arose at phase transitions and step boundary definitions.

\textbf{Conclusions:} \textit{ColoWorkflow} is the first consensus-based, validated VBA tool for comprehensive workflow analysis in minimally invasive colorectal surgery. It establishes a reproducible framework for video-based performance assessment, enabling benchmarking across institutions and supporting the development of artificial intelligence-driven workflow recognition. Its adoption may standardize training, accelerate competency acquisition, and advance data-informed surgical quality improvement.

% \keywords{Colorectal Surgery; Video Based Assessment (VBA); Artificial Intelligence (AI); Quality Improvement; Robotic Surgery; Laparoscopic Surgery}
\end{abstract}

\maketitle
\thispagestyle{firstpagestyle}

\section{Introduction}
\label{sec:introduction}

Minimally invasive surgery (MIS) has become the gold standard approach for benign and malignant elective colorectal resections due to its reduced morbidity, shorter hospital stays, and faster recovery compared with open surgery \cite{1,2}.  Despite these advantages, outcomes after MIS colorectal surgery (CRS) remain highly variable across institutions and surgeons \cite{3}. Procedural variability, difficult learning curves, and persistent complication rates underscore ongoing challenges in achieving consistent technical performance \cite{4,5,6}. Even among experienced centres and surgeons, heterogeneity in intraoperative techniques and dissection sequences contributes to variation in operative time, blood loss, complication risks, and postoperative recovery \cite{7}. Addressing this variability is critical to facilitating training, benchmarking performance, standardizing care, and in improving quality of MIS CRS.

Video-based assessment (VBA) has emerged as a robust method to evaluate and improve surgical procedures. Unlike direct observation, VBA enables asynchronous, detailed review of intraoperative workflows and critical events, supporting structured feedback and postoperative debriefing \cite{8,9}. Prior studies have demonstrated that technical performance on surgical video correlates with patient outcomes, providing an objective link between intraoperative skill and clinical results  \cite{10,11,12,13}. However, the expansion of VBA programs has been limited by the scarcity of validated assessment tools, and the significant time and expertise required for manual video review. While advances in computer vision and artificial intelligence (AI) promise to automate aspects of surgical video analysis, these computational methods still require to be trained and validated on well-defined and consistent VBA tools \cite{14}.

A standardized VBA tool for workflow analysis could quantify procedural variability in MIS CRS, identify best practices, and establish the foundation for automated workflow recognition. 

This study aimed to develop and validate a VBA tool for comprehensive workflow characterization in minimally invasive colorectal surgery. We hypothesized that a consensus-derived, procedure-agnostic framework could reliably describe workflows across laparoscopic and robotic colorectal operations and enable reproducible VBA.

\section{Materials and Methods}
\label{sec:methods}

\subsection{Study design}
\label{subsec:analysis}

This study comprised two components: (1) a modified Delphi process to establish expert consensus on workflow descriptors for minimally invasive colorectal surgery (MIS CRS), and (2) the development and validation of a video-based assessment (VBA) tool, ColoWorkflow, derived from the consensus framework. The Delphi process was conducted and reported in accordance with the ACCORD (A Consensus-Based Checklist for Reporting of Delphi Studies) guidelines \cite{15}.

\subsection{Delphi consensus on colorectal surgery workflow descriptors}
\label{subsec:delphi_consensus}

A three-round modified Delphi study was undertaken to achieve consensus on the phases and steps describing workflows in minimally invasive colorectal surgery. A multidisciplinary steering committee of seven members with expertise in colorectal surgery, VBA, and artificial intelligence (AI) oversaw the process. Expert participants were peer-nominated based on demonstrated experience in colorectal surgery and/or video-based assessment.

The steering committee developed initial workflow definitions through a comprehensive review of key surgical textbooks and literature \cite{16,17,18,19,20,21,22,23}, semi-structured interviews with colorectal surgeons, and a hierarchical task analysis (HTA) to deconstruct procedures into phases and steps. Definitions followed the SAGES consensus framework, in which \textit{phases} represent the highest-level temporal components of an operation and \textit{steps} denote procedure-specific segments achieving discrete clinical objectives \cite{24}. Preliminary workflow definitions were piloted on a multicentre dataset comprising laparoscopic and robotic CRS videos from five institutions to ensure broad procedural representation. Rigid sequencing of phases and steps was intentionally avoided to capture real-world variability in surgical practice.

The Delphi survey was implemented using JotForm (JotForm Inc., San Francisco, CA). Information on participants demographics, clinical experience, and country of practice were collected. Each proposed phase or step was rated on two 5-point Likert scales: (1) inclusion relevance, and (2) clarity and completeness of description. Participants could also provide free-text comments, suggest edits, or propose new items. Consensus thresholds were based on prior Delphi studies \cite{25}: items with $\ge$70\% agreement (Likert $\ge$4/5) were accepted, those with 60–70\% agreement or more than two revision suggestions were modified and carried forward, and items with $<$60\% agreement were excluded. Following each round, aggregated and de-identified feedback was reviewed by the steering committee to refine workflow descriptors. A summary of results was distributed to participants after each round.

All Round 2 participants were invited to the third round which consisted of an online consensus meeting. Unresolved items were reviewed with presentation of agreement data, comment counts, and item progress summaries. A steering committee moderator facilitated discussion until full consensus was achieved. Revised definitions were circulated to all participants for final review and approval.

\subsection{Development and validation of ColoWorkflow}
\label{subsec:developments_validation_coloworkflow}

Workflow descriptors reaching final consensus were incorporated into the \textit{ColoWorkflow} VBA tool. Start and end points for each phase and step were defined by the steering committee using stable, visually identifiable cues to ensure reproducible annotation. 

To assess applicability, \textit{ColoWorkflow} was used by one of the doctors in the steering committee to annotate a dataset of MIS CRS procedural videos from four European institutions and a public dataset \cite{16}.
To assess inter-rater reliability (IRR), four independent raters (a medical student, a physician, and two surgical residents) annotated a representative subset of 10 procedures, two per major procedure type. All raters had prior VBA experience and used the MOSaiC (IHU-Strasbourg, France), a web-based platform for collaborative medical video analysis \cite{26}.

\subsection{Outcomes and statistical analysis}
\label{subsec:statistical_analysis}

Participation metrics, including response and retention rates, were summarized descriptively. Mean Likert scores for each round were compared to assess convergence toward consensus. Inter-rater reliability was quantified using Cohen’s $\kappa$ and percentage agreement \cite{27}. Cohen's $\kappa$ represents the proportion of agreement beyond that expected by chance and is calculated as:
\[
\kappa = \frac{Pr(a) - Pr(e)}{1 - Pr(e)}
\]
where $Pr(a)$ denotes observed agreement and $Pr(e)$ represents expected agreement by chance. Strength of agreement was interpreted as follows: almost perfect ($\kappa > 0.90$), strong (0.80--0.90), moderate (0.60--0.79), weak (0.40--0.59), minimal (0.21--0.39), or none (0--0.20)\cite{27}. $\kappa$ values were computed for all six rater pairs and averaged.

All analyses were performed in Python (version 3.12; Python Software Foundation, Wilmington, DE) using \textit{pandas}, \textit{numpy}, and \textit{scikit-learn}. Visualizations were generated with \textit{matplotlib} and \textit{seaborn}.

\subsection{Ethical statement}
\label{subsec:ethical_statement}

Ethical approval for the collection and analysis of de-identified surgical videos was obtained under the OPERATE protocol (ID 6456). All videos were pseudonymized prior to analysis, and no identifiable patient information was included.

\section{Results}
\label{sec:results}

\subsection{Modified Delphi study for expert consensus on colorectal workflow descriptors}
\label{subsec:modified_delphi}

\begin{figure}[t]
\centering
\includegraphics[width=\columnwidth]{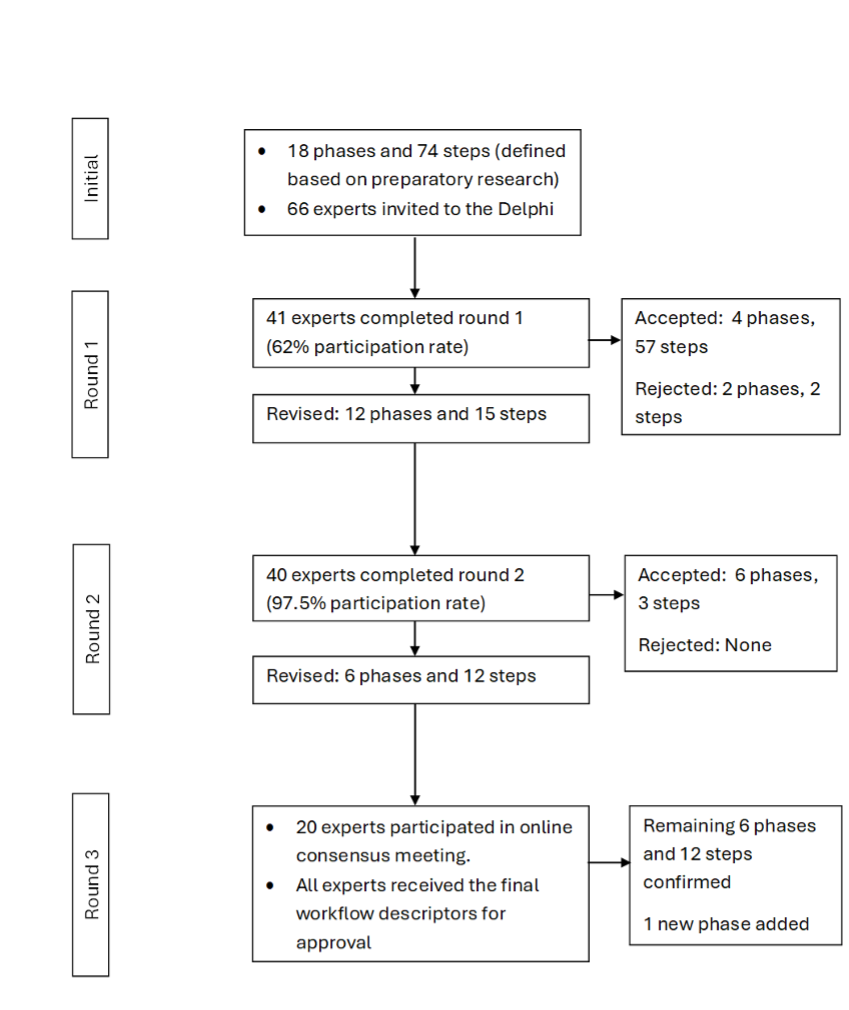}
\caption{Results of the modified Delphi process. The numbers for \textit{phases} and \textit{steps} represent both the inclusion of \textit{phases}/\textit{steps} and their corresponding descriptions as separate items for consensus.}
\label{fig:delphi_results}
\end{figure}

The modified Delphi process was conducted over three rounds spanning five months (Figure~\ref{fig:delphi_results}). Panel selection and initial item generation were completed during the first two months. Round 1 (Dec 3, 2024) remained open for 14 days, followed by Round 2 (January 3, 2025) for 16 days, with one to two reminder emails per round. The final consensus meeting (Round 3) was held online on February 27th, 2025, followed by a 10-day feedback period for final approval.

Out of 66 invited experts from 13 countries, 41 (62\%) from 11 countries completed Round 1. The panel was diverse in both geographic and procedural experience, with 40\% having performed more than 1,000 colorectal surgeries (Supplementary Table~\ref{etab:1}). Forty participants (97.5\%) continued to Round 2. The initial draft included 18 \textit{phase} items and 74 \textit{step} items describing minimally invasive (MIS) colorectal surgery (CRS) workflows. Items achieving $\ge$ 70\% agreement (Likert score $\ge$ 4/5) with minimal qualitative feedback ($\le$ 2 revision suggestions) were accepted. Average Likert scores increased between rounds for \textit{phase} items and remained stable for \textit{step} items. Most \textit{steps} (77\%) items were accepted in Round 1, with 20\% of the revised \textit{steps} subsequently accepted in Round 2. Accepted \textit{phases} items increased from 22\% to 50\% between Delphi Round 1 and 2, indicating progressive consensus among participants (Table~\ref{tab:delphi_results}). The distribution of qualitative feedback across \textit{phases} and \textit{steps} items is illustrated in Supplementary Figure~\ref{efig:1}a--b, indicating that certain procedural domains required greater refinement.

\begin{table}[t]
\centering
\caption{Results of acceptance, rejection, and revision in Rounds~1 and~2 of the Delphi. Round~3 was a collegial discussion until unanimous agreement. The numbers represent both the inclusion of \textit{phases}/\textit{steps} and their corresponding descriptions.}
\label{tab:delphi_results}
\setlength{\tabcolsep}{3pt} % tighter column spacing
\small % smaller font for compactness
\begin{tabular}{lcccc}
\toprule
\textbf{Status} & \multicolumn{2}{c}{\textbf{Round~1}} & \multicolumn{2}{c}{\textbf{Round~2}} \\
\cmidrule(lr){2-3} \cmidrule(lr){4-5}
 & \textbf{Phases (\%)} & \textbf{Steps (\%)} & \textbf{Phases (\%)} & \textbf{Steps (\%)} \\
\midrule
Accept & 4 (22.22) & 57 (77.02) & 6 (50) & 3 (20) \\
Revise & 12 (66.67) & 15 (20.27) & 6 (50) & 12 (80) \\
Reject & 2 (11.1) & 2 (2.70) & 0 & 0 \\
\bottomrule
\end{tabular}
\end{table}

\begin{figure*}[t]
\centering
\begin{minipage}[c]{0.85\textwidth}
  \includegraphics[width=\linewidth]{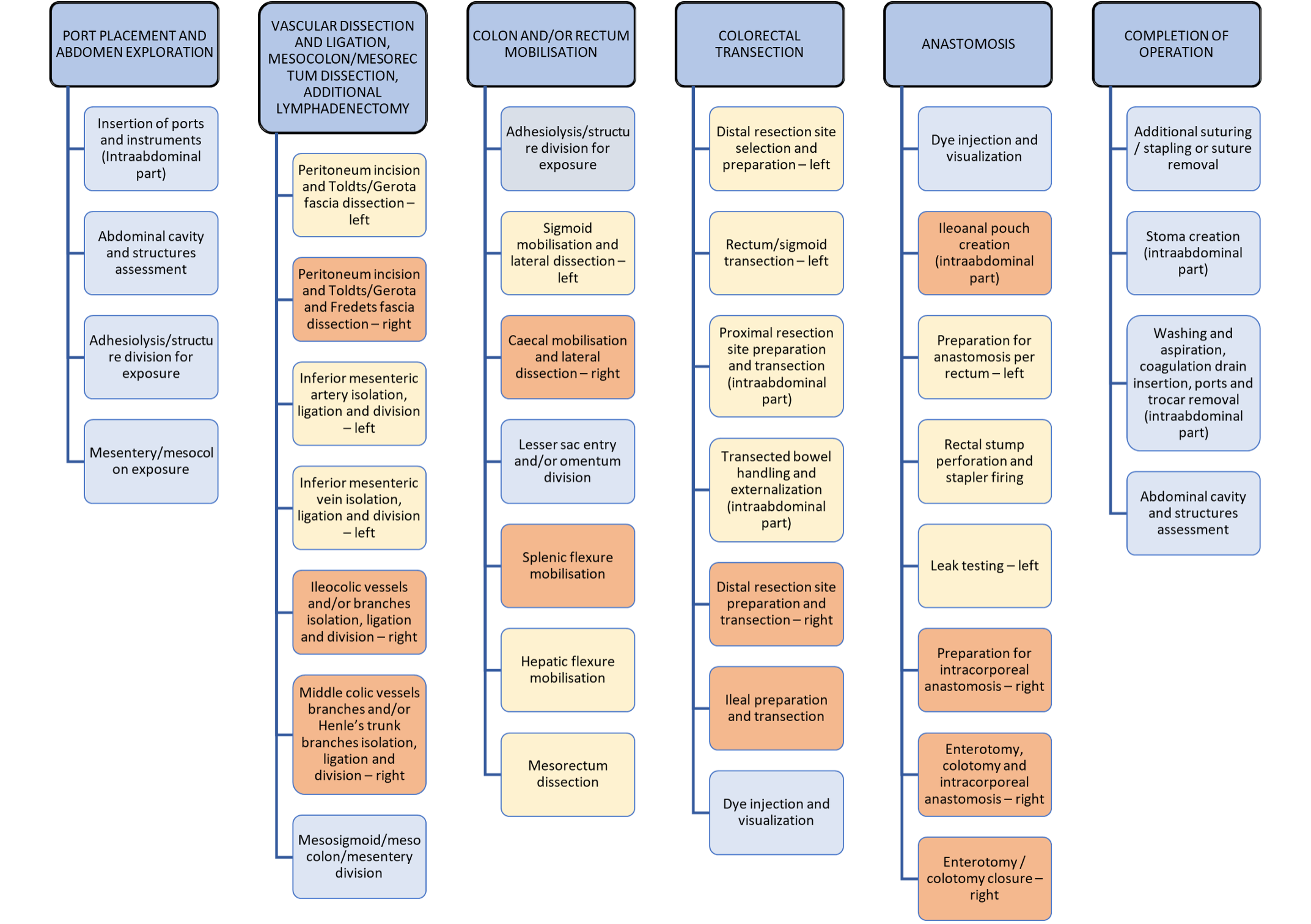}
\end{minipage}\hfill
\begin{minipage}[c]{0.15\textwidth}
  \centering
  % Increase size dramatically; adjust scale to taste (e.g., 1.4, 1.6)
  \includegraphics[angle=90,origin=c,scale=0.6]{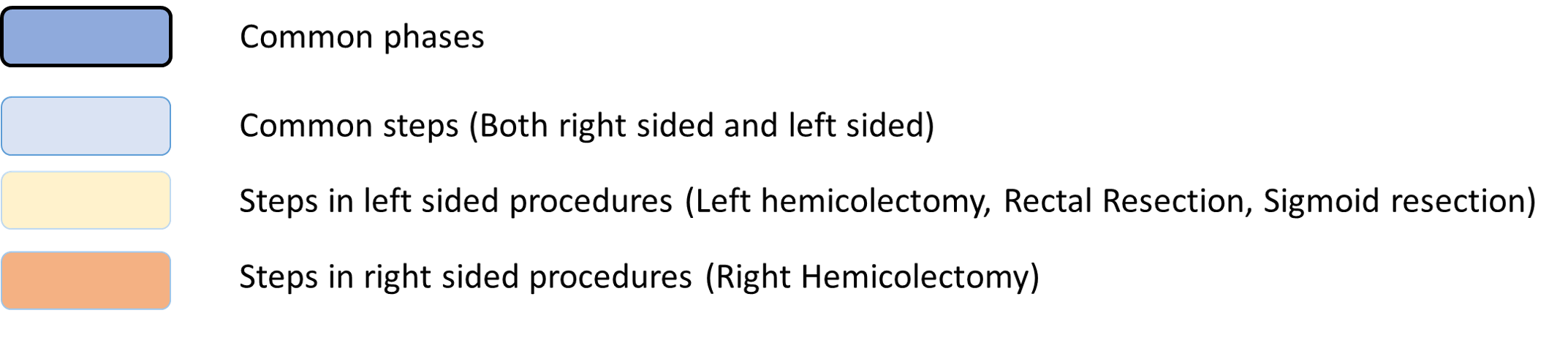}
\end{minipage}
\caption{Representation of \textit{ColoWorkflow}. Common \textit{phases} are shown in dark blue, steps shared across left- and right-sided procedures in light blue, steps specific to left-sided procedures (left hemicolectomy, rectal resection, sigmoid resection) in yellow, and steps specific to right-sided procedures (right hemicolectomy) in orange.}
\label{fig:coloworkflow_representation}
\end{figure*}

Twenty experts attended the final consensus meeting. Remaining items were reviewed, discussed, and revised for clarity and precision. Key modifications included splitting the \textit{phase} “Other Interventions” into 2 distinct \textit{phases}: “Preplanned Additional Procedures” and “Unplanned Procedures”. The definition of “Leak Testing” was modified, and the description of “Mesorectal Dissection” was refined to specify circumferential dissection with or without vascular control. 
Following these revisions, expert consensus was achieved on 10 procedural \textit{phases} and 34 workflow \textit{steps} (Figure~\ref{fig:coloworkflow_representation}) and their description. Complete \textit{phase} and \textit{step} definitions are detailed in Supplementary Tables~\ref{etab:3} and~\ref{etab:4}.

% in document

\subsection{Development and validation of ColoWorkflow}
\label{subsec:coloworkflow_development}

The video-based assessment (VBA) tool, \textit{ColoWorkflow}, was constructed using the consensus-derived workflow descriptors to ensure face and content validity according to Messick’s validity framework \cite{28}. Each \textit{phase} and \textit{step} was operationally defined with clear visual start and end cues to facilitate reproducible annotation across videos.

\textit{ColoWorkflow} was applied to 54 MIS CRS procedural videos encompassing: left hemicolectomies (10 cases), right hemicolectomies (12), rectal resections (10), sigmoid resections (12), and total proctocolectomies (10). Video durations ranged from 82 to 200 minutes with an average of $142.19 \pm 59$ minutes (see Supplementary Table~\ref{etab:2}). All \textit{phase} and \textit{step} labels except ``Ileoanal pouch creation'' were represented in at least one video, with no missing or redundant labels reported. On average, each video contained $7.52 \pm 0.79$ \textit{phases} and $16.46 \pm 2.41$ \textit{steps}. Mean \textit{phase} duration was $18.91 \pm 17.46$ minutes, and mean \textit{step} duration was $6.49 \pm 7.01$ minutes.

On a subset of 10 videos annotated by four independent raters, the mean Cohen’s $\kappa$ was 0.71 for \textit{phases} and 0.66 for \textit{steps}, indicating moderate agreement (Table~\ref{tab:irr_metrics}). Pairwise $\kappa$ values stratified by raters’ experience are reported in Table~\ref{tab:irr_metrics}. Labels with low $\kappa$ scores or high variability (highlighted in Supplementary Figure~\ref{efig:2}a--b) were examined during structured debriefings. Qualitative review revealed that disagreements stemmed primarily from inherent challenges in interpreting surgical videos, such as ambiguous visual cues or overlapping tasks, rather than inconsistent annotation practice. Raters’ feedback and insights from the qualitative review were used to refine the \textit{ColoWorkflow} VBA framework. Common pitfalls, along with recommended strategies to improve annotations, are summarized in Table~\ref{tab:irr_mitigation}.

\begin{table}[t]
\centering
\caption{Interrater reliability metrics, including Cohen’s $\kappa$ and agreement percentage.}
\label{tab:irr_metrics}
\setlength{\tabcolsep}{4pt}
\small
\resizebox{\columnwidth}{!}{
\begin{tabular}{lcccc}
\toprule
\textbf{Rater pairs} & \multicolumn{2}{c}{\textbf{Phases}} & \multicolumn{2}{c}{\textbf{Steps}} \\
\cmidrule(lr){2-3} \cmidrule(lr){4-5}
 & \textbf{Cohen’s $\kappa$} & \textbf{Agreement (\%)} & \textbf{Cohen’s $\kappa$} & \textbf{Agreement (\%)} \\
\midrule
Resident~1--Resident~2 & 0.70 & 75.74 & 0.66 & 69.41 \\
Resident~1--Med~student & 0.68 & 73.55 & 0.60 & 63.58 \\
Resident~1--Doctor & 0.74 & 78.56 & 0.73 & 75.28 \\
Resident~2--Med~student & 0.69 & 73.96 & 0.64 & 66.73 \\
Resident~2--Doctor & 0.78 & 81.89 & 0.73 & 75.09 \\
Med~student--Doctor & 0.72 & 76.48 & 0.66 & 68.47 \\
\midrule
\textbf{Average} & \textbf{0.72} & \textbf{76.70} & \textbf{0.67} & \textbf{69.76} \\
\bottomrule
\end{tabular}}
\end{table}

\begin{table*}[t]
\centering
\caption{Identified challenges contributing to interrater variability and mitigation strategies to overcome them.}
\label{tab:irr_mitigation}
\setlength{\tabcolsep}{6pt}
\small
\begin{tabular}{p{0.25\textwidth} p{0.68\textwidth}}
\toprule
\textbf{Problem} & \textbf{Mitigation strategy} \\
\midrule
Ambiguous start and end of phases and steps &
Base transitions on clear visual cues and provide illustrative examples. \\[4pt]

Hard to recognize anatomical landmarks &
Apply the VBA tool by surgeons or trained annotators supported with image primers, expert video reviews, and videos of sufficient quality. \\[4pt]

Flickering phases/steps &
Define \textit{a priori} whether to optimize for workflow stability (goal-oriented annotations, high threshold for phase/step change) or for granularity (action-oriented annotations, low threshold for phase/step change). \\[4pt]

Learning and fatigue effects &
Schedule interim calibration meetings and distribute analysis sessions over time. \\
\bottomrule
\end{tabular}
\end{table*}

\section{Discussion}
\label{sec:discussion}

This study developed and validated \textit{ColoWorkflow}, a video-based assessment (VBA) tool designed to characterize procedural workflows across minimally invasive (MIS) colorectal surgeries (CRS). Consensus on 10 procedural phases and 34 workflow steps was established through a modified Delphi process involving an international panel of experts, ensuring \textit{ColoWorkflow}'s face and content validity. Application of the tool to 54 surgical videos demonstrated its feasibility, consistency, and interrater reliability, providing preliminary evidence of response process and internal structure validity according to Messick’s framework. Collectively, these findings support \textit{ColoWorkflow} as a robust foundation for standardized workflow analysis in MIS CRS.

The present findings complement and extend prior work on VBA of CRS workflow. Previous Delphi-based frameworks have focused on individual procedures, such as laparoscopic right colectomy or sigmoid \cite{20,21,29,30}, or on discrete segments like splenic flexure mobilization \cite{31}. More recently, the Heidelberg Colorectal (HeiCo) dataset introduced a phase-based framework generalizable across left-sided CRS procedures \cite{16}. \textit{ColoWorkflow} builds on top of previous frameworks by integrating procedure-agnostic phases with procedure-specific steps within a unified hierarchical structure. It is flexible regarding the order of phases or steps to work across specific surgical approaches. This design promotes granularity where needed while enabling cross-procedure comparison, and captures observable surgical actions that are commonly encountered, regardless of institutional preferences or surgeon-specific styles.

The rigorous consensus methodology and diverse expert representation underpin the validity and generalizability of this work. The modified Delphi approach proved effective for harmonizing complex surgical descriptions. High expert engagement led to the direct acceptance of $\sim$70\% of step items in Round~1. Items with disagreement largely remained contested in Round~2 (80\%), revealing that some steps are universally agreed upon, while others reflect inherent variability in surgical technique. Disagreement often centered on terminology rather than content, emphasizing the importance of linguistic precision in workflow modeling \cite{32}. The present findings also highlight the challenge of balancing annotation granularity for clinical interpretation with the coarser segmentation needed for efficient AI model training \cite{33,34}.

The moderate ($0.6 \le \kappa \le 0.8$) interrater reliability suggests the reliability of \textit{ColoWorkflow}, providing evidence of its internal structure validity. An analysis of disagreements informed the refinement of \textit{ColoWorkflow}, integrating expert-based consensus recommendations with practical annotation insights. However, some disagreements seemed to derive from variability in mental models, VBA tool familiarity, and labeling strategies, reflecting inherent and often unavoidable differences in conceptualizing workflow definitions          \cite{35}. A structured orientation phase before any large-scale video-based analysis---including a co-review of sample videos and clearly defined usage guidelines for the VBA tool---could help overcome these challenges.

Altogether, several aspects distinguish \textit{ColoWorkflow} from other frameworks for workflow analysis. Its thorough development through a highly participated modified Delphi exercise and its application to a multicenter dataset spanning CRS procedures provide early but convincing evidence of the tool’s validity. \textit{ColoWorkflow}'s hierarchical and flexible structure allows analysis across procedure types, surgical platforms (laparoscopic and robotic), and institutional protocols; such generalizability should favor applicability, cross-center benchmarking, and continuous professional development. Focusing on observable actions rather than prescriptive sequencing, it reflects real-world diversity while maintaining analytic rigor, facilitating VBA research. Importantly, the tool’s explicit visual definitions and modular structure facilitate the development of AI for CRS workflow analysis, providing a bridge between expert consensus and machine learning scalability.

This work has limitations. First, the collegial discussion during the final Delphi round may have reduced independent expression due to group dynamics. Nonetheless, such discussion followed two independent consensus rounds and was of critical importance to guarantee convergence on workflow descriptors, probably contributing to \textit{ColoWorkflow}'s face and content validity. Despite broad international participation, the expert panel may not have fully captured regional or technique-specific variations in surgical practice. Additionally, while the multicenter dataset included multiple CRS procedures, some were not represented (e.g., pouch creation). Future studies will easily integrate more steps or procedural variations into \textit{ColoWorkflow} thanks to its hierarchical design and flexibility regarding phase and step sequencing. Finally, raters did not include expert CRS surgeons. While including more experienced raters might have improved interrater reliability, it was preferred to enroll medical students and residents trained on VBA, as workflow analysis does not entail assessments of high surgical semantics (e.g., analyzing anastomosis quality). Considering how overburdened expert surgeons are, a tool trainees can use should facilitate implementation.

Despite these limitations, this work provides a critical foundation for future research and clinical translation. Establishing a validated, generalizable colorectal workflow schema enables more consistent measurement of intraoperative performance and facilitates cross-institutional comparisons. When combined with advances in computer vision and AI, \textit{ColoWorkflow} can accelerate the development of automated systems for workflow recognition, skill assessment, and intraoperative guidance. Moreover, standardized workflow mapping can enhance training curricula and quality improvement initiatives by linking intraoperative processes to patient outcomes.

In conclusion, \textit{ColoWorkflow} represents the first expert-validated, procedure-generalizable framework for analyzing minimally invasive colorectal surgical workflows. Through a rigorous consensus and validation process, it establishes a shared language for procedural analysis that is both granular and adaptable. Future work will focus on expanding validity evidence, integrating additional procedure types, and leveraging \textit{ColoWorkflow} as a foundation for AI-based workflow analysis and surgical education.

\section{Disclosures}

Pietro Mascagni and Nicolas Padoy are co-founders and shareholders of Scialytics. Jim Khan and Eloy Espin-Basany serve as proctors for Intuitive Surgical. Isacco Montroni has received compensation for participation in courses organized by Olympus and Fresenius Kabi. Monica Ortenzi is a paid collaborator with Theator. Willem A. Bemelman is a shareholder in Semiflex. All other authors declare no conflicts of interest.

\section{Acknowledgement}

This study was not preregistered in an independent institutional registry. De-identified consensus data and analytical methods data may be made available from the corresponding author on reasonable request.

\subsection{Data Availability Statement}

De-identified video data may be made available from the corresponding author on reasonable request, pending institutional approvals and data-use agreements.

\subsection{Funding Statement}

This work has received funding from the European Union (ERC, CompSURG, 101088553). Views and opinions expressed are, however, those of the authors only and do not necessarily reflect those of the European Union or the European Research Council. Neither the European Union nor the granting authority can be held responsible for them. This work was also developed within the Interdisciplinary Thematic Institute HealthTech (ITI 2021-2028 program of the University of Strasbourg, CNRS and Inserm), supported by IdEx Unistra (ANR-10-IDEX-0002) and SFRI (STRATUS project, ANR-20-SFRI-0012) under the framework of the French Investments for the Future Program. This work was partially supported by French state funds managed by the ANR under Grant ANR-10-IAHU-02 (IHU Strasbourg).

\section*{CompSurg Colorectal Experts Group Collaborators}

{\normalsize
Alberto Arezzo\textsuperscript{a}, Alfonso Lapergola\textsuperscript{b}, Amjad Parvaiz\textsuperscript{c}, Deena Harji\textsuperscript{d}, Domenico D’Ugo\textsuperscript{e}, Elisa Cassinotti\textsuperscript{f}, Eloy Espin-Basany\textsuperscript{g}, Emanuele Rausa\textsuperscript{h}, Francesco Brucchi\textsuperscript{i}, Gabriele Anania\textsuperscript{j}, Gaya Spolverato\textsuperscript{k}, Gianluca Rizzo\textsuperscript{l}, Giovanni Guglielmo Laracca\textsuperscript{m}, Giuseppe Quero\textsuperscript{n}, Graziano Pernazza\textsuperscript{o}, Isacco Montroni\textsuperscript{p}, Jared Torkington\textsuperscript{q}, Jim Khan\textsuperscript{r}, Manish Chand\textsuperscript{s}, Maria Vannucci\textsuperscript{t}, Michel Adamina\textsuperscript{u}, Monica Ortenzi\textsuperscript{v}, Nariaki Okamoto\textsuperscript{w}, Paola De Nardi\textsuperscript{x}, Paolo Pietro Bianchi\textsuperscript{y}, Patricia Sylla\textsuperscript{z}, Ronan Cahill\textsuperscript{aa}, Stefano Rausei\textsuperscript{ab}, Umberto Bracale\textsuperscript{ac}, Vincenzo Tondolo\textsuperscript{ad}, Willem A.~Bemelman\textsuperscript{ae}

\vspace{0.8em}
{\small\textit{
\begin{itemize}
  \item[a] Department of Surgical Sciences, University of Turin, Turin, Italy
  \item[b] Visceral and Digestive Surgery Department, Colorectal and Robotic Unit, Nouvel Hôpital Civil, University Hospitals of Strasbourg, Strasbourg, France; IRCAD (Research Institute against Digestive Cancer), Strasbourg, France; IHU (Institut Hospitalo-Universitaire), Strasbourg, France
  \item[c] Champalimaud Foundation, Lisbon, Portugal
  \item[d] Department of Colorectal Surgery, Manchester University NHS Foundation Trust, Manchester, UK; Clinical Trials Research Unit, University of Leeds, Leeds, UK
  \item[e] Fondazione Policlinico Universitario Gemelli -- IRCCS, Rome, Italy
  \item[f] Department of General and Minimally Invasive Surgery, Fondazione IRCCS Ca’ Granda Ospedale Maggiore Policlinico, Milan, Italy; Department of Clinical Sciences and Community Health, University of Milan, Milan, Italy
  \item[g] General Surgery Department, Hospital Vall d’Hebron, Universitat Autònoma de Barcelona, Spain
  \item[h] Colorectal Surgery Unit, Fondazione IRCCS Istituto Nazionale dei Tumori, Milan, Italy
  \item[i] University of Milan, Milan, Italy
  \item[j] Department of Medical Sciences, University of Ferrara, Ferrara, Italy
  \item[k] General Surgery~3, Azienda Ospedale Università di Padova; Department of Surgical Oncological and Gastrointestinal Sciences, University of Padova, Padova, Italy
  \item[l] UOC Chirurgia Digestiva e del Colon-Retto, Ospedale Isola Tiberina Gemelli Isola, Rome, Italy
  \item[m] AOU Sant’Andrea, Surgery Department, Rome, Italy
  \item[n] Fondazione Policlinico Universitario Agostino Gemelli IRCCS di Roma; Università Cattolica del Sacro Cuore di Roma, Italy
  \item[o] San Giovanni Addolorata Hospital, Department of Surgery, Rome, Italy
  \item[p] IRCCS Fondazione Istituto Nazionale dei Tumori, Division of Colon and Rectal Cancer Surgery, Milan, Italy
  \item[q] University Hospital of Wales, Cardiff, UK
  \item[r] Portsmouth Hospitals University NHS Trust, Portsmouth, UK; University of Portsmouth, UK
  \item[s] University College London, London, UK
  \item[t] University of Torino, Department of General Surgery, Torino, Italy; Institute of Image-Guided Surgery, IHU-Strasbourg, Strasbourg, France
  \item[u] Department of Visceral Surgery and Medicine, Inselspital, Bern University Hospital, Bern, Switzerland
  \item[v] Università Politecnica delle Marche, Ancona, Italy
  \item[w] IRCAD (Research Institute against Digestive Cancer), Strasbourg, France; Department of Colorectal Surgery, National Cancer Center Hospital East, Kashiwa, Japan
  \item[x] Colorectal Surgery, IRCCS San Raffaele Scientific Institute, Milan, Italy
  \item[y] ASST Santi Paolo e Carlo, Università di Milano, Milan, Italy
  \item[z] Mount Sinai Hospital, New York, NY, USA
  \item[aa] UCD Centre for Precision Surgery, University College Dublin, Dublin, Ireland; Department of Surgery, Mater Misericordiae University Hospital, Dublin, Ireland
  \item[ab] Department of Surgery, ASST Sette Laghi, Cittiglio--Angera (Varese), Italy
  \item[ac] Department of Medicine, Surgery and Odontoiatry, University of Salerno, Salerno, Italy
  \item[ad] UOC Chirurgia Digestiva e Colon-Retto, Ospedale Isola Tiberina--Gemelli Isola, Rome, Italy; Università Cattolica del Sacro Cuore, Rome, Italy
  \item[ae] Department of Surgery, Amsterdam University Medical Centers, Amsterdam, The Netherlands
\end{itemize}
}}
}

\bibliographystyle{elsarticle-num}
\bibliography{arxiv}

@article{24,
	title = {{SAGES} consensus recommendations on an annotation framework for surgical video},
	volume = {35},
	issn = {1432-2218},
	url = {https://doi.org/10.1007/s00464-021-08578-9},
	doi = {10.1007/s00464-021-08578-9},
	language = {en},
	number = {9},
	urldate = {2025-06-04},
	journal = {Surgical Endoscopy},
	author = {Meireles, Ozanan R. and Rosman, Guy and Altieri, Maria S. and Carin, Lawrence and Hager, Gregory and Madani, Amin and Padoy, Nicolas and Pugh, Carla M. and Sylla, Patricia and Ward, Thomas M. and Hashimoto, Daniel A. and {the SAGES Video Annotation for AI Working Groups}},
	month = sep,
	year = {2021},
	pages = {4918--4929},
}

@article{32,
	title = {Deep learning in surgical process modeling: {A} systematic review of workflow recognition},
	volume = {162},
	issn = {1532-0464},
	shorttitle = {Deep learning in surgical process modeling},
	url = {https://www.sciencedirect.com/science/article/pii/S1532046425000085},
	doi = {10.1016/j.jbi.2025.104779},
	urldate = {2025-06-04},
	journal = {Journal of Biomedical Informatics},
	author = {Liu, Zhenzhong and Chen, Kelong and Wang, Shuai and Xiao, Yijun and Zhang, Guobin},
	month = feb,
	year = {2025},
	pages = {104779},
}

@article{34,
	title = {Computer {Vision} {Analysis} of {Intraoperative} {Video}: {Automated} {Recognition} of {Operative} {Steps} in {Laparoscopic} {Sleeve} {Gastrectomy}},
	volume = {270},
	issn = {0003-4932},
	shorttitle = {Computer {Vision} {Analysis} of {Intraoperative} {Video}},
	url = {https://journals.lww.com/annalsofsurgery/abstract/2019/09000/computer_vision_analysis_of_intraoperative_video_.3.aspx},
	doi = {10.1097/SLA.0000000000003460},
	language = {en-US},
	number = {3},
	urldate = {2025-06-04},
	journal = {Annals of Surgery},
	author = {Hashimoto, Daniel A. and Rosman, Guy and Witkowski, Elan R. and Stafford, Caitlin and Navarette-Welton, Allison J. and Rattner, David W. and Lillemoe, Keith D. and Rus, Daniela L. and Meireles, Ozanan R.},
	month = sep,
	year = {2019},
	pages = {414},
}

@inproceedings{33,
	title = {Recovery of {Surgical} {Workflow}: a {Model}-based {Approach}},
	shorttitle = {Recovery of {Surgical} {Workflow}},
	booktitle = {21st {International} {Congress} and {Exhibition} on {Computer} {Assisted} {Radiology} and {Surgery}},
	author = {Padoy, Nicolas and Horn, Martin and Feussner, Hubertus and marie-odile, Berger and Navab, Nassir},
	month = jun,
	year = {2007},
}

@article{16,
	title = {Heidelberg colorectal data set for surgical data science in the sensor operating room},
	volume = {8},
	copyright = {2021 The Author(s)},
	issn = {2052-4463},
	url = {https://www.nature.com/articles/s41597-021-00882-2},
	doi = {10.1038/s41597-021-00882-2},
	language = {en},
	number = {1},
	urldate = {2025-06-04},
	journal = {Scientific Data},
	author = {Maier-Hein, Lena and Wagner, Martin and Ross, Tobias and Reinke, Annika and Bodenstedt, Sebastian and Full, Peter M. and Hempe, Hellena and Mindroc-Filimon, Diana and Scholz, Patrick and Tran, Thuy Nuong and Bruno, Pierangela and Kisilenko, Anna and Müller, Benjamin and Davitashvili, Tornike and Capek, Manuela and Tizabi, Minu D. and Eisenmann, Matthias and Adler, Tim J. and Gröhl, Janek and Schellenberg, Melanie and Seidlitz, Silvia and Lai, T. Y. Emmy and Pekdemir, Bünyamin and Roethlingshoefer, Veith and Both, Fabian and Bittel, Sebastian and Mengler, Marc and Mündermann, Lars and Apitz, Martin and Kopp-Schneider, Annette and Speidel, Stefanie and Nickel, Felix and Probst, Pascal and Kenngott, Hannes G. and Müller-Stich, Beat P.},
	month = apr,
	year = {2021},
	note = {Publisher: Nature Publishing Group},
	pages = {101},
}

@article{20,
	title = {Procedural key steps in laparoscopic colorectal surgery, consensus through {Delphi} methodology},
	volume = {29},
	issn = {1432-2218},
	url = {https://doi.org/10.1007/s00464-014-3979-7},
	doi = {10.1007/s00464-014-3979-7},
	language = {en},
	number = {9},
	urldate = {2025-06-04},
	journal = {Surgical Endoscopy},
	author = {Dijkstra, Frederieke A. and Bosker, Robbert J. I. and Veeger, Nicolaas J. G. M. and van Det, Marc J. and Pierie, Jean Pierre E. N.},
	month = sep,
	year = {2015},
	pages = {2620--2627},
}

@article{22,
	title = {Real-time automatic surgical phase recognition in laparoscopic sigmoidectomy using the convolutional neural network-based deep learning approach},
	volume = {34},
	issn = {1432-2218},
	url = {https://doi.org/10.1007/s00464-019-07281-0},
	doi = {10.1007/s00464-019-07281-0},
	language = {en},
	number = {11},
	urldate = {2025-06-04},
	journal = {Surgical Endoscopy},
	author = {Kitaguchi, Daichi and Takeshita, Nobuyoshi and Matsuzaki, Hiroki and Takano, Hiroaki and Owada, Yohei and Enomoto, Tsuyoshi and Oda, Tatsuya and Miura, Hirohisa and Yamanashi, Takahiro and Watanabe, Masahiko and Sato, Daisuke and Sugomori, Yusuke and Hara, Seigo and Ito, Masaaki},
	month = nov,
	year = {2020},
	pages = {4924--4931},
}

@article{21,
	title = {Development of an objective evaluation tool to assess technical skill in laparoscopic colorectal surgery: a {Delphi} methodology},
	volume = {201},
	issn = {0002-9610},
	shorttitle = {Development of an objective evaluation tool to assess technical skill in laparoscopic colorectal surgery},
	url = {https://www.sciencedirect.com/science/article/pii/S0002961010002606},
	doi = {10.1016/j.amjsurg.2010.01.031},
	number = {2},
	urldate = {2025-06-04},
	journal = {The American Journal of Surgery},
	author = {Palter, Vanessa N. and MacRae, Helen M. and Grantcharov, Teodor P.},
	month = feb,
	year = {2011},
	pages = {251--259},
}

@article{35,
	title = {Challenges in surgical video annotation},
	volume = {26},
	issn = {null},
	url = {https://doi.org/10.1080/24699322.2021.1937320},
	doi = {10.1080/24699322.2021.1937320},
	number = {1},
	urldate = {2025-06-04},
	journal = {Computer Assisted Surgery},
	author = {Ward, Thomas M. and , Danyal M., Fer and , Yutong, Ban and , Guy, Rosman and , Ozanan R., Meireles and and Hashimoto, Daniel A.},
	month = jan,
	year = {2021},
	pmid = {34126014},
	note = {Publisher: Taylor \& Francis
\_eprint: https://doi.org/10.1080/24699322.2021.1937320},
	pages = {58--68},
}

@article{10,
	title = {The association between video-based assessment of intraoperative technical performance and patient outcomes: a systematic review},
	volume = {36},
	issn = {1432-2218},
	shorttitle = {The association between video-based assessment of intraoperative technical performance and patient outcomes},
	url = {https://doi.org/10.1007/s00464-022-09296-6},
	doi = {10.1007/s00464-022-09296-6},
	language = {en},
	number = {11},
	urldate = {2025-06-04},
	journal = {Surgical Endoscopy},
	author = {Balvardi, Saba and Kammili, Anitha and Hanson, Melissa and Mueller, Carmen and Vassiliou, Melina and Lee, Lawrence and Schwartzman, Kevin and Fiore, Julio F. and Feldman, Liane S.},
	month = nov,
	year = {2022},
	pages = {7938--7948},
}

@article{9,
	title = {The what? {How}? {And} {Who}? {Of} video based assessment},
	volume = {221},
	issn = {0002-9610},
	shorttitle = {The what?},
	url = {https://www.sciencedirect.com/science/article/pii/S0002961020303809},
	doi = {10.1016/j.amjsurg.2020.06.027},
	number = {1},
	urldate = {2025-06-04},
	journal = {The American Journal of Surgery},
	author = {Pugh, Carla M. and Hashimoto, Daniel A. and Korndorffer, James R.},
	month = jan,
	year = {2021},
	pages = {13--18},
}

@article{8,
	title = {Video-{Based} {Assessment} of {Surgical} {Quality}—{Will} {Video} {Kill} the {Radio} {Star}?},
	volume = {7},
	issn = {2574-3805},
	url = {https://doi.org/10.1001/jamanetworkopen.2024.6477},
	doi = {10.1001/jamanetworkopen.2024.6477},
	number = {4},
	urldate = {2025-06-04},
	journal = {JAMA Network Open},
	author = {Reitz, Alexandra C. W. and Massarweh, Nader N.},
	month = apr,
	year = {2024},
	pages = {e246477},
}

@book{18,
	title = {Colon and {Rectal} {Surgery}: {Anorectal} {Operations}},
	isbn = {978-1-4511-5490-0},
	shorttitle = {Colon and {Rectal} {Surgery}},
	language = {en},
	publisher = {Lippincott Williams \& Wilkins},
	author = {Wexner, Steven D. and Fleshman, James W.},
	month = jan,
	year = {2012},
}

@book{19,
	title = {Colon and {Rectal} {Surgery}},
	isbn = {978-0-7817-4043-2},
	language = {en},
	publisher = {Lippincott Williams \& Wilkins},
	author = {Corman, Marvin L.},
	year = {2005},
	note = {Google-Books-ID: imEv3PqkJSEC},
}

@book{17,
	title = {Farquharson's {Textbook} of {Operative} {General} {Surgery}},
	isbn = {978-1-4441-7593-6},
	language = {en},
	publisher = {CRC Press},
	author = {Farquharson, Margaret and Hollingshead, James and Moran, Brendan},
	month = oct,
	year = {2014},
	note = {Google-Books-ID: JtEgCAAAQBAJ},
}

@article{27,
	title = {Interrater reliability: the kappa statistic},
	volume = {22},
	issn = {1330-0962, 1846-7482},
	shorttitle = {Interrater reliability},
	url = {https://hrcak.srce.hr/89395},
	language = {en},
	number = {3},
	urldate = {2025-06-04},
	journal = {Biochemia Medica},
	author = {McHugh, Mary L.},
	month = oct,
	year = {2012},
	note = {Publisher: Hrvatsko društvo za medicinsku biokemiju i laboratorijsku medicinu},
	pages = {276--282},
}

@article{15,
	title = {{ACCORD} ({ACcurate} {COnsensus} {Reporting} {Document}): {A} reporting guideline for consensus methods in biomedicine developed via a modified {Delphi}},
	volume = {21},
	issn = {1549-1676},
	shorttitle = {{ACCORD} ({ACcurate} {COnsensus} {Reporting} {Document})},
	url = {https://journals.plos.org/plosmedicine/article?id=10.1371/journal.pmed.1004326},
	doi = {10.1371/journal.pmed.1004326},
	language = {en},
	number = {1},
	urldate = {2025-06-04},
	journal = {PLOS Medicine},
	author = {Gattrell, William T. and Logullo, Patricia and Zuuren, Esther J. van and Price, Amy and Hughes, Ellen L. and Blazey, Paul and Winchester, Christopher C. and Tovey, David and Goldman, Keith and Hungin, Amrit Pali and Harrison, Niall},
	month = jan,
	year = {2024},
	note = {Publisher: Public Library of Science},
	pages = {e1004326},
}

@article{23,
	title = {Laparoscopic right hemicolectomy with {CME}: standardization using the “critical view” concept},
	volume = {32},
	issn = {1432-2218},
	shorttitle = {Laparoscopic right hemicolectomy with {CME}},
	url = {https://doi.org/10.1007/s00464-018-6267-0},
	doi = {10.1007/s00464-018-6267-0},
	language = {en},
	number = {12},
	urldate = {2025-06-04},
	journal = {Surgical Endoscopy},
	author = {Strey, Christoph Werner and Wullstein, Christoph and Adamina, Michel and Agha, Ayman and Aselmann, Heiko and Becker, Thomas and Grützmann, Robert and Kneist, Werner and Maak, Matthias and Mann, Benno and Moesta, Kurt Thomas and Runkel, Norbert and Schafmayer, Clemens and Türler, Andreas and Wedel, Thilo and Benz, Stefan},
	month = dec,
	year = {2018},
	pages = {5021--5030},
}

@misc{26,
	title = {{MOSaiC}: a {Web}-based {Platform} for {Collaborative} {Medical} {Video} {Assessment} and {Annotation}},
	shorttitle = {{MOSaiC}},
	doi = {10.48550/arXiv.2312.08593},
	publisher = {arXiv},
	author = {Mazellier, Jean-Paul and Boujon, Antoine and Bour-Lang, Méline and Erharhd, Maël and Waechter, Julien and Wernert, Emilie and Mascagni, Pietro and Padoy, Nicolas},
	note = {arXiv:2312.08593 [cs]},
}

@article{31,
	title = {Three {Surgical} {Approaches} of {Laparoscopic} {Splenic} {Flexure} {Mobilization}},
	volume = {22},
	issn = {2234-778X, 2234-5248},
	url = {http://www.e-jmis.org/journal/view.html?doi=10.7602/jmis.2019.22.2.85},
	doi = {10.7602/jmis.2019.22.2.85},
	language = {en},
	number = {2},
	urldate = {2025-06-05},
	journal = {The Journal of Minimally Invasive Surgery},
	author = {Lee, Yoon Suk},
	month = jun,
	year = {2019},
	pages = {85--86},
}

@article{12,
	title = {Association {Between} {Surgeon} {Technical} {Skills} and {Patient} {Outcomes}},
	volume = {155},
	issn = {2168-6254},
	url = {https://www.ncbi.nlm.nih.gov/pmc/articles/PMC7439214/},
	doi = {10.1001/jamasurg.2020.3007},
	number = {10},
	urldate = {2025-06-21},
	journal = {JAMA Surgery},
	author = {Stulberg, Jonah J. and Huang, Reiping and Kreutzer, Lindsey and Ban, Kristen and Champagne, Bradley J. and Steele, Scott R. and Johnson, Julie K. and Holl, Jane L. and Greenberg, Caprice C. and Bilimoria, Karl Y.},
	month = oct,
	year = {2020},
	pmid = {32838425},
	pmcid = {PMC7439214},
	pages = {960--968},
}

@article{11,
	title = {Association {Between} {Surgical} {Technical} {Skill} and {Long}-term {Survival} for {Colon} {Cancer}},
	volume = {7},
	issn = {2374-2437},
	url = {https://www.ncbi.nlm.nih.gov/pmc/articles/PMC7600051/},
	doi = {10.1001/jamaoncol.2020.5462},
	number = {1},
	urldate = {2025-06-21},
	journal = {JAMA Oncology},
	author = {Brajcich, Brian C. and Stulberg, Jonah J. and Palis, Bryan E. and Chung, Jeanette W. and Huang, Reiping and Nelson, Heidi and Bilimoria, Karl Y.},
	month = jan,
	year = {2021},
	pmid = {33125472},
	pmcid = {PMC7600051},
	pages = {127--129},
}

@article{13,
	title = {Evaluating the {Effect} of {Surgical} {Skill} on {Outcomes} for {Laparoscopic} {Sleeve} {Gastrectomy}: {A} {Video}-based {Study}},
	volume = {273},
	issn = {0003-4932},
	shorttitle = {Evaluating the {Effect} of {Surgical} {Skill} on {Outcomes} for {Laparoscopic} {Sleeve} {Gastrectomy}},
	url = {https://journals.lww.com/annalsofsurgery/abstract/2021/04000/evaluating_the_effect_of_surgical_skill_on.21.aspx},
	doi = {10.1097/SLA.0000000000003385},
	language = {en-US},
	number = {4},
	urldate = {2025-06-21},
	journal = {Annals of Surgery},
	author = {Varban, Oliver A. and Thumma, Jyothi R. and Finks, Jonathan F. and Carlin, Arthur M. and Ghaferi, Amir A. and Dimick, Justin B.},
	month = apr,
	year = {2021},
	pages = {766},
}

@article{14,
	title = {Computer vision in surgery: from potential to clinical value},
	volume = {5},
	copyright = {2022 The Author(s)},
	issn = {2398-6352},
	shorttitle = {Computer vision in surgery},
	url = {https://www.nature.com/articles/s41746-022-00707-5},
	doi = {10.1038/s41746-022-00707-5},
	language = {en},
	number = {1},
	urldate = {2025-09-06},
	journal = {npj Digital Medicine},
	author = {Mascagni, Pietro and Alapatt, Deepak and Sestini, Luca and Altieri, Maria S. and Madani, Amin and Watanabe, Yusuke and Alseidi, Adnan and Redan, Jay A. and Alfieri, Sergio and Costamagna, Guido and Boškoski, Ivo and Padoy, Nicolas and Hashimoto, Daniel A.},
	month = oct,
	year = {2022},
	note = {Publisher: Nature Publishing Group},
	pages = {163},
}

@article{28,
	title = {Standards of {Validity} and the {Validity} of {Standards} in {Performance} {Asessment}},
	volume = {14},
	issn = {1745-3992},
	url = {https://onlinelibrary.wiley.com/doi/abs/10.1111/j.1745-3992.1995.tb00881.x},
	doi = {10.1111/j.1745-3992.1995.tb00881.x},
	language = {en},
	number = {4},
	urldate = {2025-10-08},
	journal = {Educational Measurement: Issues and Practice},
	author = {Messick, Samuel},
	year = {1995},
	note = {\_eprint: https://onlinelibrary.wiley.com/doi/pdf/10.1111/j.1745-3992.1995.tb00881.x},
	pages = {5--8},
}

@article{3,
	title = {Laparoscopic {Colorectal} {Surgery} {Outcomes} {Improved} {After} {National} {Training} {Program} ({LAPCO}) for {Specialists} in {England}},
	volume = {275},
	issn = {0003-4932},
	url = {https://journals.lww.com/annalsofsurgery/abstract/2022/06000/laparoscopic_colorectal_surgery_outcomes_improved.19.aspx},
	doi = {10.1097/SLA.0000000000004584},
	language = {en-US},
	number = {6},
	urldate = {2025-10-08},
	journal = {Annals of Surgery},
	author = {Hanna, George B. and Mackenzie, Hugh and Miskovic, Danilo and Ni, Melody and Wyles, Susannah and Aylin, Paul and Parvaiz, Amjad and Cecil, Tom and Gudgeon, Andrew and Griffith, John and Robinson, Jonathan M. and Selvasekar, Chelliah and Rockall, Tim and Acheson, Austin and Maxwell-Armstrong, Charles and Jenkins, John T. and Horgan, Alan and Cunningham, Chris and Lindsey, Ian and Arulampalam, Tan and Motson, Roger W. and Francis, Nader K. and Kennedy, Robin H. and Coleman, Mark G.},
	month = jun,
	year = {2022},
	pages = {1149},
}

@article{29,
	title = {Standardization of the surgical technique and reporting for radical right colectomy with central vascular ligation and complete mesocolic excision ({RRoC}-{STAR}): {Delphi} consensus},
	volume = {9},
	issn = {2474-9842},
	shorttitle = {Standardization of the surgical technique and reporting for radical right colectomy with central vascular ligation and complete mesocolic excision ({RRoC}-{STAR})},
	url = {https://doi.org/10.1093/bjsopen/zraf066},
	doi = {10.1093/bjsopen/zraf066},
	number = {3},
	urldate = {2025-10-22},
	journal = {BJS Open},
	author = {Sica, Giuseppe S and Anania, Gabriele and Fiorani, Cristina and Siragusa, Leandro and Vinci, Danilo and Caricato, Marco and Delrio, Paolo and Agrusa, Antonino and Baldazzi, Gianandrea and Reddavid, Rossella and Pellino, Gianluca and {RRoC-STAR Collaborative Group}},
	month = jun,
	year = {2025},
	pages = {zraf066},
}

@article{4,
	title = {Impact of variability among surgeons on postoperative morbidity and mortality and ultimate survival.},
	volume = {302},
	issn = {0959-8138, 1468-5833},
	url = {https://www.bmj.com/content/302/6791/1501},
	doi = {10.1136/bmj.302.6791.1501},
	language = {en},
	number = {6791},
	urldate = {2025-10-22},
	journal = {British Medical Journal},
	author = {McArdle, C. S. and Hole, D.},
	month = jun,
	year = {1991},
	pmid = {1713087},
	note = {Publisher: British Medical Journal Publishing Group
Section: Research Article},
	pages = {1501--1505},
}

@article{5,
	title = {The {Learning} {Curve} for {Laparoscopic} {Colorectal} {Surgery}: {Preliminary} {Results} {From} a {Prospective} {Analysis} of 1194 {Laparoscopic}-{Assisted} {Colectomies}},
	volume = {132},
	issn = {0004-0010},
	shorttitle = {The {Learning} {Curve} for {Laparoscopic} {Colorectal} {Surgery}},
	url = {https://doi.org/10.1001/archsurg.1997.01430250043009},
	doi = {10.1001/archsurg.1997.01430250043009},
	number = {1},
	urldate = {2025-10-22},
	journal = {Archives of Surgery},
	author = {Bennett, Charles L. and Stryker, Steven J. and Ferreira, M. Rosario and Adams, John and Beart, Jr, Robert W.},
	month = jan,
	year = {1997},
	pages = {41--44},
}

@article{7,
	title = {Association of {Surgical} {Skill} {Assessment} {With} {Clinical} {Outcomes} in {Cancer} {Surgery}},
	volume = {155},
	issn = {2168-6262},
	doi = {10.1001/jamasurg.2020.1004},
	language = {eng},
	number = {7},
	journal = {JAMA surgery},
	author = {Curtis, Nathan J. and Foster, Jake D. and Miskovic, Danilo and Brown, Chris S. B. and Hewett, Peter J. and Abbott, Sarah and Hanna, George B. and Stevenson, Andrew R. L. and Francis, Nader K.},
	month = jul,
	year = {2020},
	pmid = {32374371},
	pmcid = {PMC7203671},
	pages = {590--598},
}

@article{25,
	title = {International e-{Delphi} survey to define best practice in the reporting of intracranial pressure monitoring recording data},
	volume = {4},
	issn = {2772-5294},
	url = {https://www.sciencedirect.com/science/article/pii/S2772529424001164},
	doi = {10.1016/j.bas.2024.102860},
	urldate = {2025-10-22},
	journal = {Brain and Spine},
	author = {Kommer, Maya and Hawthorne, Christopher and Moss, Laura and Piper, Ian and O'Kane, Roddy and Czosnyka, Marek and Enblad, Per and Hemphill, J Claude and Spiegelberg, Andreas and Riddell, John S. and Shaw, Martin},
	month = jan,
	year = {2024},
	pages = {102860},
}

@article{1,
	title = {Clinical {Practice} {Guidelines} for {Enhanced} {Recovery} {After} {Colon} and {Rectal} {Surgery} {From} the {American} {Society} of {Colon} and {Rectal} {Surgeons} and {Society} of {American} {Gastrointestinal} and {Endoscopic} {Surgeons}},
	volume = {60},
	issn = {0012-3706},
	url = {https://journals.lww.com/dcrjournal/fulltext/2017/08000/clinical_practice_guidelines_for_enhanced_recovery.3.aspx},
	doi = {10.1097/DCR.0000000000000883},
	language = {en-US},
	number = {8},
	urldate = {2025-10-29},
	journal = {Diseases of the Colon \& Rectum},
	author = {Carmichael, Joseph C. and Keller, Deborah S. and Baldini, Gabriele and Bordeianou, Liliana and Weiss, Eric and Lee, Lawrence and Boutros, Marylise and McClane, James and Feldman, Liane S. and Steele, Scott R.},
	month = aug,
	year = {2017},
	pages = {761},
}

@article{2,
	title = {Enhanced {Recovery} {Program} in {Colorectal} {Surgery}: {A} {Meta}-analysis of {Randomized} {Controlled} {Trials}},
	volume = {38},
	copyright = {© 2014 The Author(s) under exclusive licence to Société Internationale de Chirurgie},
	issn = {1432-2323},
	shorttitle = {Enhanced {Recovery} {Program} in {Colorectal} {Surgery}},
	url = {https://onlinelibrary.wiley.com/doi/abs/10.1007/s00268-013-2416-8},
	doi = {10.1007/s00268-013-2416-8},
	language = {en},
	number = {6},
	urldate = {2025-10-29},
	journal = {World Journal of Surgery},
	author = {Greco, Massimiliano and Capretti, Giovanni and Beretta, Luigi and Gemma, Marco and Pecorelli, Nicolò and Braga, Marco},
	year = {2014},
	note = {\_eprint: https://onlinelibrary.wiley.com/doi/pdf/10.1007/s00268-013-2416-8},
	pages = {1},
}

@article{6,
	title = {Initiating statistical process control to improve quality outcomes in colorectal surgery},
	volume = {29},
	issn = {1432-2218},
	url = {https://doi.org/10.1007/s00464-015-4108-y},
	doi = {10.1007/s00464-015-4108-y},
	language = {en},
	number = {12},
	urldate = {2025-10-29},
	journal = {Surgical Endoscopy},
	author = {Keller, Deborah S. and Stulberg, Jonah J. and Lawrence, Justin K. and Samia, Hoda and Delaney, Conor P.},
	month = dec,
	year = {2015},
	pages = {3559--3564},
}

@article{30,
	title = {Achieving high quality standards in laparoscopic colon resection for cancer: {A} {Delphi} consensus-based position paper},
	volume = {44},
	issn = {07487983},
	shorttitle = {Achieving high quality standards in laparoscopic colon resection for cancer},
	url = {https://linkinghub.elsevier.com/retrieve/pii/S0748798318301288},
	doi = {10.1016/j.ejso.2018.01.091},
	language = {en},
	number = {4},
	urldate = {2025-11-01},
	journal = {European Journal of Surgical Oncology},
	author = {Lorenzon, Laura and Biondi, Alberto and Carus, Thomas and Dziki, Adam and Espin, Eloy and Figueiredo, Nuno and Ruiz, Marcos Gomez and Mersich, Tamas and Montroni, Isacco and Tanis, Pieter J. and Benz, Stefan Rolf and Bianchi, Paolo Pietro and Biebl, Matthias and Broeders, Ivo and De Luca, Raffaele and Delrio, Paolo and D'Hondt, Mathieu and Fürst, Alois and Grosek, Jan and Guimaraes Videira, Jose Flavio and Herbst, Friedrich and Jayne, David and Lázár, György and Miskovic, Danilo and Muratore, Andrea and Helmer Sjo, Ole and Scheinin, Tom and Tomazic, Ales and Türler, Andreas and Van De Velde, Cornelius and Wexner, Steven D. and Wullstein, Christoph and Zegarski, Wojciech and D'Ugo, Domenico},
	month = apr,
	year = {2018},
	pages = {469--483},
}

% optional: switch numbering to eTables for the supplement
% ---------- shared settings ----------

\clearpage
\setcounter{table}{0}
\setcounter{figure}{0}
\captionsetup[table]{labelformat=simple, labelsep=period, name={eTable}}
\captionsetup[figure]{labelformat=simple, labelsep=period, name={eFigure}}

{\LARGE \bfseries Supplementary Materials}\par\vspace{3em}
% ============================================================
% eTable 1 — Demographics of participants
% ============================================================

\begin{table}[h]
\onecolumn
\centering
\caption{Demographics of participants of the Delphi study.}
\label{etab:1}
\setlength{\tabcolsep}{6pt}
\small

\onecolumn
\begin{adjustbox}{center}
\begin{tabular}{l l}
\toprule
\textbf{Variable} & \textbf{N (\%)} \\
\midrule
\multicolumn{2}{l}{\textbf{Age (years)}} \\[2pt]
26--30 & 4 (9.8) \\
31--35 & 2 (4.9) \\
36--40 & 9 (22.0) \\
41--45 & 5 (12.2) \\
46--50 & 6 (14.6) \\
51--55 & 4 (9.8) \\
$>$55  & 11 (26.8) \\[4pt]

\multicolumn{2}{l}{\textbf{Highest professional qualification}} \\[2pt]
Head of Department & 10 (24.4) \\
Professor & 16 (39.0) \\
Consultant & 9 (22.0) \\
Fellow & 2 (4.9) \\
Resident year~4 & 2 (4.9) \\
Resident year~5 & 2 (4.9) \\[4pt]

\multicolumn{2}{l}{\textbf{Total number of colorectal surgeries performed}} \\[2pt]
1--200 & 8 (19.5) \\
201--400 & 9 (22.0) \\
401--600 & 2 (4.9) \\
601--800 & 4 (9.8) \\
801--1000 & 2 (4.9) \\
$>$1000 & 16 (39.0) \\[4pt]

\multicolumn{2}{l}{\textbf{Number of colorectal surgeries performed per year}} \\[2pt]
1--50 & 13 (31.7) \\
51--100 & 12 (29.3) \\
101--200 & 3 (7.3) \\
201--300 & 12 (29.3) \\
$>$300 & 1 (2.4) \\
\bottomrule
\end{tabular}
\end{adjustbox}

\end{table}

\vspace{1em}

% ============================================================
% eTable 2 — Dataset Description
% ============================================================

\begin{table}[h!]
\centering
\onecolumn
\caption{Description of the dataset used for validation of \textit{ColoWorkflow}.}
\twocolumn
\label{etab:2}
\setlength{\tabcolsep}{8pt}
\small
\begin{tabular}{l l c c}
\toprule
\textbf{Procedure type} & \textbf{Source of data} & \textbf{Number of videos} & \textbf{Average duration (min)} \\
\midrule
Left Hemicolectomy & Center~1, Center~2, Center~3 & 10 & 100.82 \\
Right Hemicolectomy & Center~1, Center~2, Center~3 & 12 & 93.00 \\
Rectal Resection & Center~2, Center~3, HeiCo dataset & 10 & 184.85 \\
Sigmoidectomy & Center~2, Center~3, Center~4, HeiCo dataset & 12 & 134.31 \\
Total Proctocolectomy & HeiCo dataset & 10 & 209.32 \\
\bottomrule
\end{tabular}
\end{table}

\vspace{1em}

% ============================================================
% eFigure 1 — Delphi Phases and Steps
% ============================================================

\begin{figure*}[t]
\centering
\includegraphics[width=0.8\textwidth]{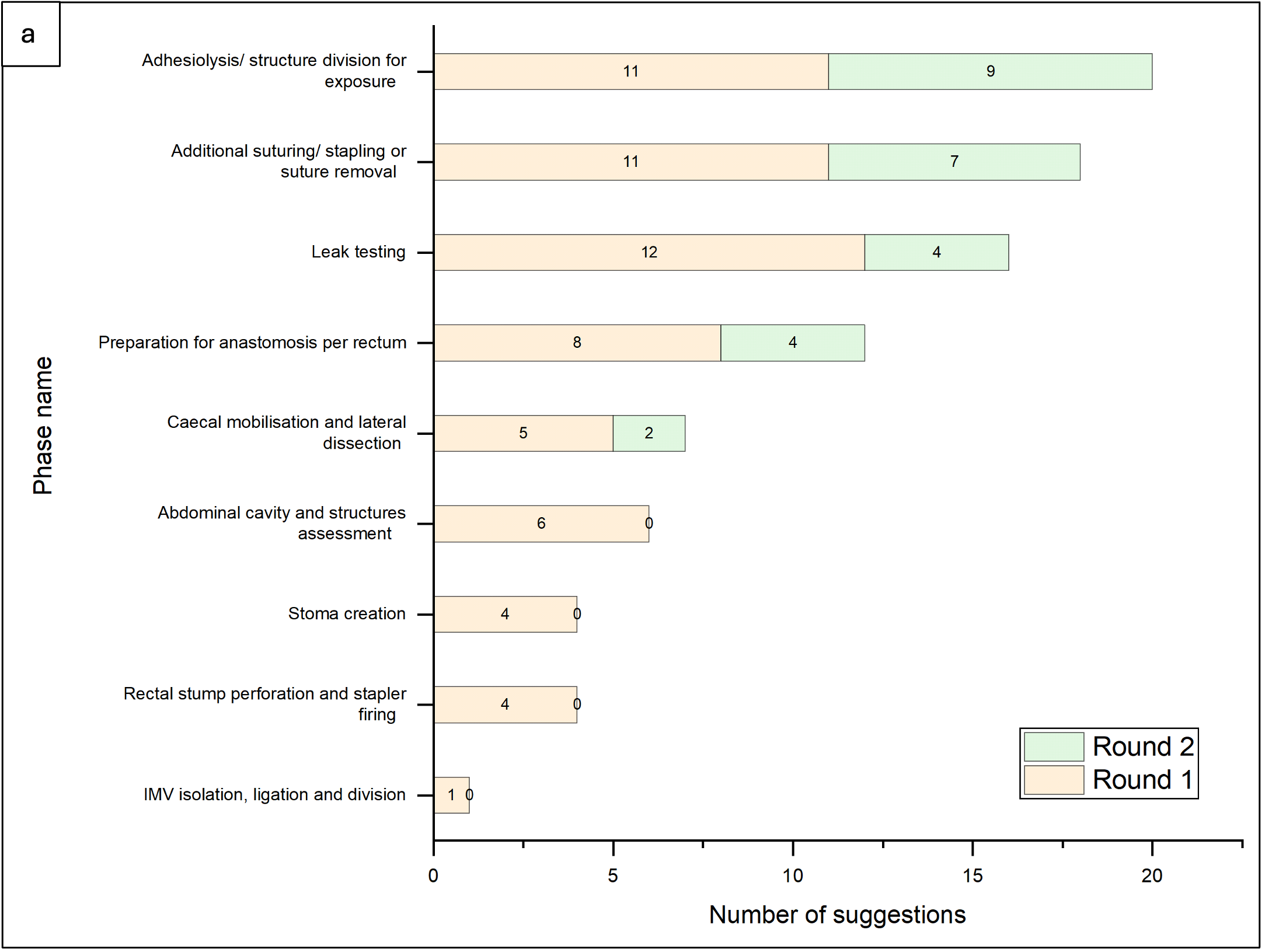}\vspace{4pt}
\includegraphics[width=0.8\textwidth]{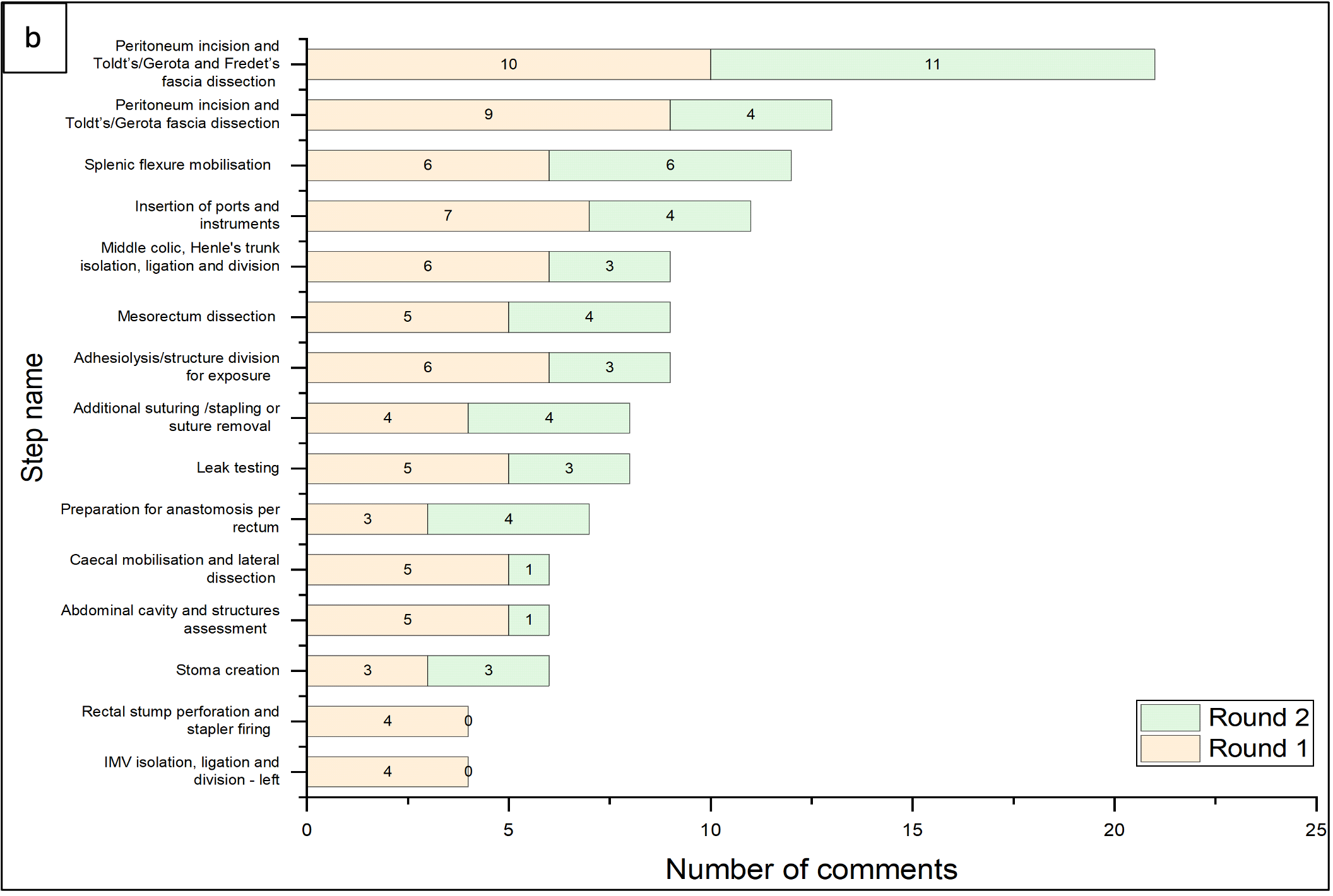}
\onecolumn
\caption{Phases and steps in the video-based assessment (VBA) tool ranked by number of expert suggestions given for each phase/step naming and description. All phases shown in (a) (above); Top~15 steps shown in (b) (below). Comments mostly revolved around alternate approaches, inclusion of anatomical landmarks, and clarity in wording.}
\twocolumn
\label{efig:1}
\end{figure*}

\vspace{1em}

% ============================================================
% eTable 3 — Phases
% ============================================================

\begin{table*}[p]
\centering
\onecolumn
\caption{List of phases and their descriptions reaching consensus after all rounds of Delphi. Numbering is for reference only and does not necessarily reflect the required order of phases and steps.}
\twocolumn
\label{etab:3}
\setlength{\tabcolsep}{3pt}
\small
\begin{tabularx}{\textwidth}{@{} c p{0.25\textwidth} p{0.70\textwidth} @{}}
\toprule
\textbf{No.} & \textbf{Phase name} & \textbf{Description} \\
\midrule
1 & Port placement and abdomen exploration & Laparoscopic camera insertion, additional port placement, cavity exploration, adhesiolysis, and manipulation of tissues/organs to expose the surgical field before starting executive phases. \\
2 & Vascular dissection and ligation, mesocolon/mesorectum dissection, additional lymphadenectomy & Identification, dissection, and ligation of vascular pedicles to achieve procedure-specific lymphadenectomy, and dissection of the mesentery (mesocolon or mesorectum). \\
3 & Colon and/or rectum mobilisation & Dissection of lateral attachments of the colon (ascending, descending, sigmoid), with/without continued splenic or hepatic flexure mobilisation. \\
4 & Colorectal transection & Dissection and transection of proximal and distal bowel margins, with or without perfusion assessment. \\
5 & Anastomosis & Reestablishment of bowel continuity. \\
6 & Completion of operation & Final inspection of the abdominal cavity (hemostasis, viability, orientation, foreign body check), with/without mesenteric defect closure, drain placement, stoma creation (if planned), specimen retrieval, and port removal. \\
7 & Preplanned additional procedures & Interventions not routinely part of colorectal surgery but preplanned (e.g., concurrent organ work, additional resections). \\
8 & Unplanned procedures & Intraoperative deviations from planned workflow due to complications or judgment-based decisions. \\
9 & Extracorporeal procedures & Dye visualization of the extracted colon for vascular assessment prior to proximal resection; may include ileoanal pouch construction. \\
\bottomrule
\end{tabularx}
\end{table*}

\vspace{2em}

% ============================================================
% eTable 4 — Steps
% ============================================================

\onecolumn
\setlength{\tabcolsep}{3pt}
\small
\setlength{\LTleft}{0pt}
\setlength{\LTright}{0pt}

\begin{longtable}{@{} c p{0.28\textwidth} p{0.68\textwidth} @{}}
\caption{Video-based assessment tool: list of steps and their descriptions reaching consensus after all rounds of Delphi. Numbering is for reference only and does not indicate sequential order.}
\label{etab:4} \\
\toprule
\textbf{No.} & \textbf{Step name} & \textbf{Description} \\
\midrule
\endfirsthead

\caption[]{Video-based assessment tool: list of steps and their descriptions reaching consensus after all rounds of Delphi. (continued)} \\
\toprule
\textbf{No.} & \textbf{Step name} & \textbf{Description} \\
\midrule
\endhead

\midrule
\multicolumn{3}{r}{\textit{Continued on next page}} \\
\endfoot

\bottomrule
\endlastfoot

1 & Insertion of ports and instruments (intraabdominal part) & Trocar entry into the abdominal cavity and eventual insertion of laparoscopic instruments. \\
2 & Abdominal cavity and structures assessment & Visual and instrumental assessment of bowel, liver, peritoneum, and other intraabdominal structures. \\
3 & Adhesiolysis/structure division for exposure & Division of adhesions and/or falciform ligament to delineate anatomy and optimize exposure. \\
4 & Mesentery/mesocolon exposure & Lifting and repositioning omentum, small bowel, or colon to expose the mesentery or mesocolon root. \\
5 & Peritoneum incision and Toldt’s/Gerota fascia dissection -- left & Peritoneal incision with dissection to isolate inferior mesenteric vessels and develop the retroperitoneal plane between Toldt and Gerota fasciae. \\
6 & Peritoneum incision and Toldt’s/Gerota/Fredet fascia dissection -- right & Peritoneal incision with retroperitoneal dissection to isolate relevant vessels, visualize the duodenum, and develop planes between Toldt, Gerota, and Fredet fasciae. \\
7 & Inferior mesenteric artery isolation, ligation and division -- left & Dissection, ligation, and division of the inferior mesenteric artery. \\
8 & Inferior mesenteric vein isolation, ligation and division -- left & Identification, dissection, ligation, and division of the inferior mesenteric vein near the ligament of Treitz. \\
9 & Ileocolic vessels isolation, ligation and division -- right & Dissection, ligation, and division of the ileocolic artery and vein. \\
10 & Middle colic and/or Henle trunk branches isolation, ligation and division -- right & Dissection, ligation, and division of middle colic branches with or without right colic artery and Henle trunk branches. \\
11 & Mesosigmoid/mesocolon/mesentery division & Division of mesosigmoid, mesocolon, or mesentery parallel to bowel wall or along vascular pedicles in preparation for bowel transection. \\
12 & Sigmoid mobilisation and lateral dissection -- left & Lateral mobilization of sigmoid and descending colon by dividing the lateral peritoneal reflection along the Monk line. \\
13 & Caecal mobilisation and lateral dissection -- right & Lateral mobilization of caecum and ascending colon by dividing inferior and lateral peritoneal attachments along the Monk line. \\
14 & Lesser sac entry and/or omentum division & Division of gastrocolic ligament or colo-epiploic attachments to enter the lesser sac and/or divide omentum. \\
15 & Splenic flexure mobilisation & Division of splenocolic, pancreaticocolic, and phrenocolic ligaments with mobilization of the splenic flexure. \\
16 & Hepatic flexure mobilisation & Division of hepatic flexure attachments to mobilize the right colonic flexure. \\
17 & Mesorectum dissection & Circumferential dissection of the mesorectum with or without vascular control down to the planned transection level. \\
18 & Distal resection site selection and preparation -- left & Circumferential clearing of pericolonic or perirectal fat at the planned distal resection site with optional perfusion assessment. \\
19 & Rectum/sigmoid transection -- left & Transection of the distal bowel segment at the planned resection level using stapling devices. \\
20 & Proximal resection site preparation and transection (intraabdominal part) & Intraabdominal preparation and marking of the proximal bowel with circumferential clearing for planned transection. \\
21 & Transected bowel handling and externalization (intraabdominal part) & Intraabdominal preparation for bowel externalization including incision, wound protector placement, and manipulation. \\
22 & Distal resection site preparation and transection -- right & Preparation and transection of the distal bowel end for right-sided resections with optional perfusion assessment. \\
23 & Ileal preparation and transection & Preparation and transection of the proximal ileal segment for right-sided resections with optional perfusion assessment. \\
24 & Dye injection and visualization & Injection of dye to assess bowel perfusion and guide resection margins. \\
25 & Ileoanal pouch creation (intraabdominal part) & Construction of an ileal pouch for ileoanal anastomosis with attachment of the stapler anvil. \\
26 & Preparation for anastomosis per rectum -- left & Advancement and positioning of the proximal bowel into the pelvis for tension-free colorectal anastomosis. \\
27 & Rectal stump perforation and stapler firing & Perforation of rectal stump with stapler spike followed by stapler approximation and firing. \\
28 & Leak testing -- left & Assessment of anastomotic integrity using air, saline, dye, and/or endoscopic evaluation. \\
29 & Preparation for intracorporeal anastomosis -- right & Alignment and positioning of proximal and distal bowel ends for intracorporeal anastomosis. \\
30 & Enterotomy, colotomy and intracorporeal anastomosis -- right & Creation of enterotomy and colotomy followed by stapled intracorporeal anastomosis and bleeding assessment. \\
31 & Enterotomy/colotomy closure -- right & Closure of the common enterotomy to restore bowel continuity. \\
32 & Additional suturing, stapling or suture removal & Reinforcement of anastomosis, closure of mesenteric defects, and/or removal of temporary sutures. \\
33 & Stoma creation (intraabdominal part) & Preparation and positioning of bowel for stoma creation with verification of orientation and tension. \\
34 & Washing, aspiration, drain insertion, and port removal (intraabdominal part) & Irrigation, aspiration, hemostasis, drain placement if required, and removal of trocars and ports. \\

\end{longtable}

\twocolumn
\vspace{2em}

% ============================================================
% eFigure 2 — Inter-annotator variability
% ============================================================

\begin{figure*}[t]
\centering
\includegraphics[width=\textwidth]{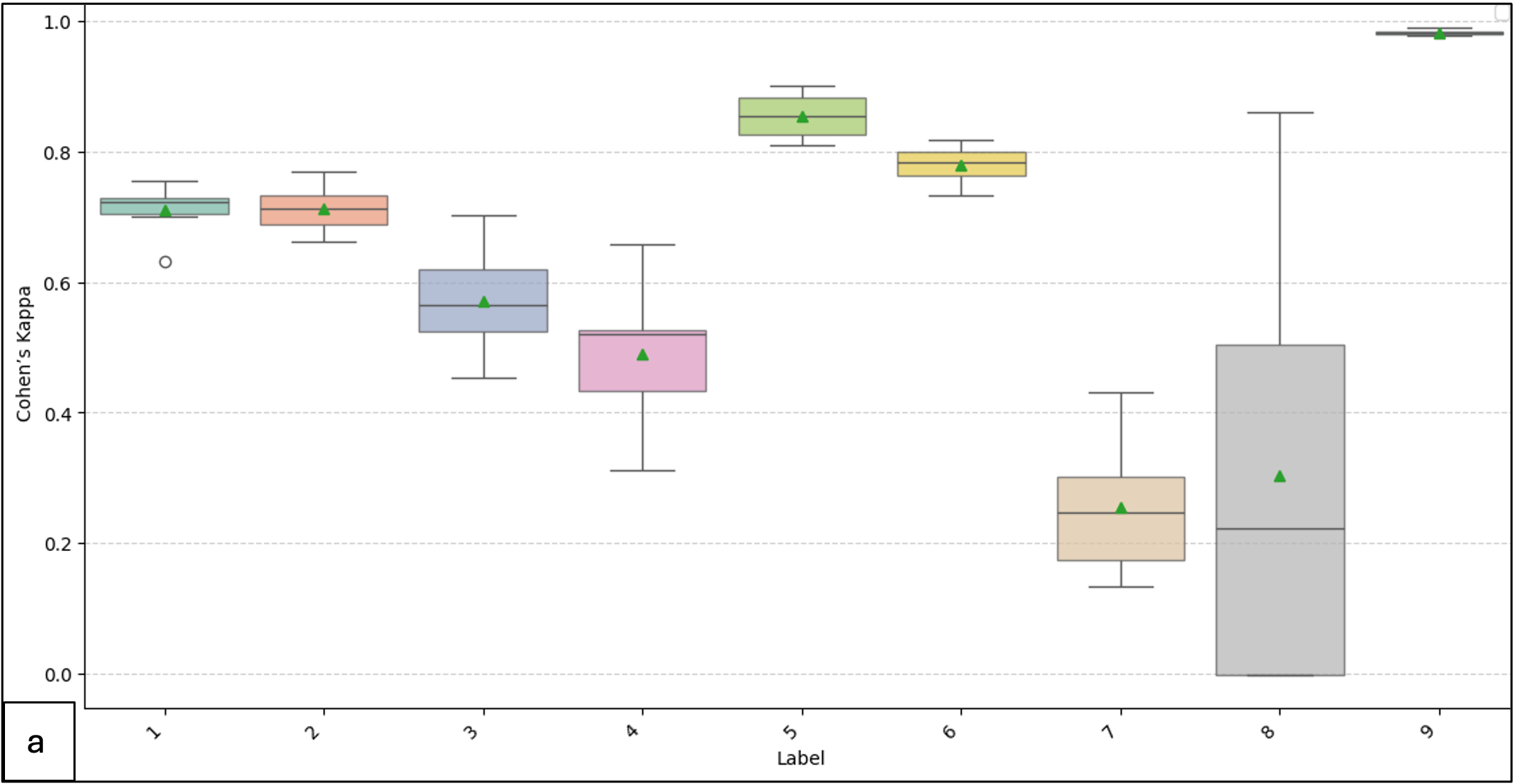}\vspace{4pt}
\includegraphics[width=\textwidth]{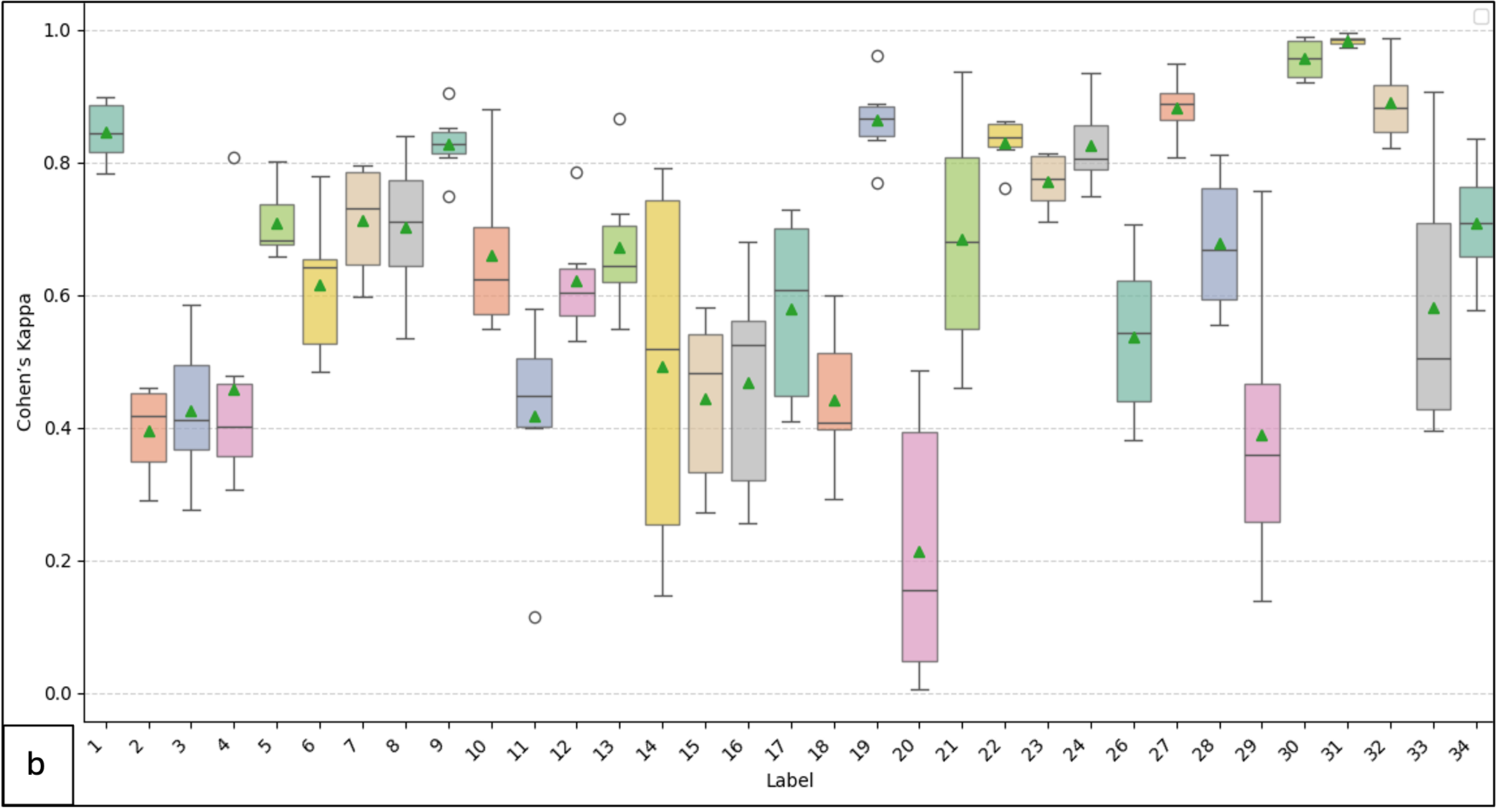}
\caption{Inter-annotator differences between the annotators and distribution across the labels. The plots show the Cohen’s $\kappa$ metric and its variability for each label, indicating that some labels have high agreement with low variation (e.g., step~1), while others may have low agreement and high variability (e.g., step~20). Phases shown in (a) (above); Steps in (b) (below). Phase and step names correspond to the numbering given in eTables~\ref{etab:3} and~\ref{etab:4}.}
\label{efig:2}
\end{figure*}

\end{document}